\documentclass{article}

    \PassOptionsToPackage{square,numbers}{natbib}

\usepackage[final]{neurips_2019}

\usepackage[utf8]{inputenc} 
\usepackage[T1]{fontenc}    
\usepackage{hyperref}       
\usepackage{url}            
\usepackage{booktabs}       
\usepackage{amsfonts}       
\usepackage{nicefrac}       
\usepackage{microtype}      
\usepackage{graphicx}
\usepackage{mathtools}

\usepackage[dvipsnames,usenames]{xcolor}
\newcommand{\dam}[1]{\textbf{\textcolor{blue}{[[DAM: #1]]}}}

\usepackage{subcaption}
\usepackage{varwidth}

\title{Probabilistic Bias Mitigation in Word Embeddings}

\author{%
  Hailey James \\
  Harvard College\\
  Cambridge, MA\\
  \texttt{haileyjames@college.harvard.edu} \\
   \And
   David Alvarez-Melis \\
   Massachusetts Institute of Technology \\
   Cambridge, MA \\
   \texttt{dalvmel@mit.edu} \\
}

\begin{document}

\maketitle
\begin{abstract}
It has been shown that word embeddings derived from large corpora tend to incorporate biases present in their training data. Various methods for mitigating these biases have been proposed, but recent work has demonstrated that these methods hide but fail to truly remove the biases, which can still be observed in word nearest-neighbor statistics. In this work we propose a probabilistic view of word embedding bias. We leverage this framework to present a novel method for mitigating bias which relies on probabilistic observations to yield a more robust bias mitigation algorithm. We demonstrate that this method effectively reduces bias according to three separate measures of bias while maintaining embedding quality across various popular benchmark semantic tasks.
\end{abstract}

\section{Introduction}

Word embeddings, or vector representations of words, are an important component of Natural Language Processing (NLP) models and necessary for many downstream tasks. However, word embeddings, including embeddings commonly deployed for public use, have been shown to exhibit unwanted societal stereotypes and biases, raising concerns about disparate impact on axes of gender, race, ethnicity, and religion \citep{DBLP:journals/corr/BolukbasiCZSK16a, DBLP:journals/corr/IslamBN16}. The impact of this bias has manifested in a range of downstream tasks, ranging from autocomplete suggestions \cite{googleautocomplete} to advertisement delivery \cite{lambrecht}, increasing the likelihood of amplifying harmful biases through the use of these models.

The most well-established method thus far for mitigating bias\footnote{We intentionally do not reference the resulting embeddings as "debiased" or free from all gender bias, and prefer the term "mitigating bias" rather that "debiasing," to guard against the misconception that the resulting embeddings are entirely "safe" and need not be critically evaluated for bias in downstream tasks.} relies on projecting target words\footnote{Throughout this paper, we use \textit{word} interchangeably with the vector representing the word in an embedding.} onto a bias subspace (such as a gender subspace) and subtracting out the difference between the resulting distances \cite{DBLP:journals/corr/BolukbasiCZSK16a}. On the other hand, the most popular metric for measuring bias is the WEAT statistic \cite{DBLP:journals/corr/IslamBN16}, which compares the cosine similarities between groups of words. However, WEAT has been recently shown to overestimate bias as a result of implicitly relying on similar frequencies for the target words \cite{Ethayarajh2019UnderstandingUW}, and \citet{DBLP:journals/corr/abs-1903-03862} demonstrated that evidence of bias can still be recovered after geometric bias mitigation by examining the neighborhood of a target word among socially-biased words.

In response to this, we propose an alternative framework for bias mitigation in word embeddings that approaches this problem from a probabilistic perspective. The motivation for this approach is two-fold. First, most popular word embedding algorithms are probabilistic at their core -- i.e., they are trained (explicitly or implicitly \citep{levy2014neural}) to minimize some form of word co-occurrence probabilities. Thus, we argue that a framework for measuring and treating bias in these embeddings should take into account, in addition to their geometric aspect, their probabilistic nature too. On the other hand, the issue of bias has also been approached (albeit in different contexts) in the fairness literature, where various intuitive notions of equity such as equalized odds have been formalized through probabilistic criteria. By considering analogous criteria for the word embedding setting, we seek to draw connections between these two bodies of work. 

We present experiments on various bias mitigation benchmarks and show that our framework is comparable to state-of-the-art alternatives according to measures of geometric bias mitigation and that it performs far better according to measures of neighborhood bias. For fair comparison, we focus on mitigating a binary gender bias in pre-trained word embeddings using SGNS (skip-gram with negative-sampling), though we note that this framework and methods could be extended to other types of bias and word embedding algorithms.

\section{Background}

\paragraph{Geometric Bias Mitigation}

Geometric bias mitigation uses the cosine distances between words to both measure and remove gender bias \cite{DBLP:journals/corr/BolukbasiCZSK16a}. This method implicitly defines bias as a geometric asymmetry between words when projected onto a subspace, such as the gender subspace constructed from a set of gender pairs such as $\mathcal{P} = \{(he,she),(man,woman),(king,queen)...\}$. The projection of a vector $v$ onto $B$ (the subspace) is defined by $v_B = \sum_{j=1}^{k} (v \cdot b_j) b_j$ where a subspace $B$ is
defined by k orthogonal unit vectors $B = {b_1,...,b_k}$.



\subparagraph{WEAT}

The WEAT statistic \cite{DBLP:journals/corr/IslamBN16} demonstrates the presence of biases in word embeddings with an effect size defined as the mean test statistic across the two word sets:

\begin{equation}
\frac{mean_{x \in X} s(x,A,B) - mean_{y \in Y} s(y,A,B)}{std\_ dev_{w \in X \cup Y} s(w,A,B)}
\end{equation}

Where $s$, the test statistic, is defined as: $s(w,A,B) = mean_{a \in A} cos(w,a) - mean_{b \in B} cos(w,a)$, and $X$,$Y$,$A$, and $B$ are groups of words for which the association is measured. Possible values range from $-2$ to $2$ depending on the association of the words groups, and a value of zero indicates $X$ and $Y$ are equally associated with $A$ and $B$. See \citet{Ethayarajh2019UnderstandingUW} for further details on WEAT.



\subparagraph{RIPA}

The RIPA (relational inner product association) metric was developed as an alternative to WEAT, with the critique that WEAT is likely to overestimate the bias of a target attribute \cite{Ethayarajh2019UnderstandingUW}. The RIPA metric formalizes the measure of bias used in geometric bias mitigation as the inner product association of a word vector $v$ with respect to a relation vector $b$. The relation vector is constructed from the first principal component of the differences between gender word pairs. We report the absolute value of the RIPA metric as the value can be positive or negative according to the direction of the bias. A value of zero indicates a lack of bias, and the value is bound by $[-||w||,||w||]$.

\subparagraph{Neighborhood Metric}
The neighborhood bias metric proposed by \citet{DBLP:journals/corr/abs-1903-03862} quantifies bias as the proportion of male socially-biased words among the $k$ nearest socially-biased male and female neighboring words, whereby biased words are obtained by projecting neutral words onto a gender relation vector. As we only examine the target word among the 1000 most socially-biased words in the vocabulary (500 male and 500 female), a word’s bias is measured as the ratio of its neighborhood of socially-biased male and socially-biased female words, so that  a value of 0.5 in this metric would indicate a perfectly unbiased word, and values closer to 0 and 1 indicate stronger bias. 




\section{A Probabilistic Framework for Bias Mitigation}

Our objective here is to extend and complement the geometric notions of word embedding bias described in the previous section with an alternative, probabilistic, approach. Intuitively, we seek a notion of \textit{equality} akin to that of \textit{demographic parity} in the fairness literature, which requires that a decision or outcome be independent of a protected attribute such as gender. \cite{DBLP:journals/corr/HardtPS16}. Similarly, when considering a probabilistic definition of unbiased in word embeddings, we can consider the conditional probabilities of word pairs, ensuring for example that $p(doctor|man) \approx p(doctor|woman)$, and can extend this probabilistic framework to include the neighborhood of a target word, addressing the potential pitfalls of geometric bias mitigation.

Conveniently, most word embedding frameworks allow for immediate computation of the conditional probabilities $P(w|c)$. Here, we focus our attention on the Skip-Gram method with Negative Sampling (SGNS) of \citet{DBLP:journals/corr/MikolovSCCD13}, although our framework can be equivalently instantiated for most other popular embedding methods, owing to their core similarities \citep{levy2014neural, hashimoto2016word}. Leveraging this probabilistic nature, we construct a bias mitigation method in two steps, and examine each step as an independent method as well as the resulting composite method.

\paragraph{Probabilistic Bias Mitigation}

This component of our bias mitigation framework seeks to enforce that the probability of prediction or outcome cannot depend on a protected class such as gender. We can formalize this intuitive goal through a loss function that penalizes the discrepancy between the conditional probabilities of a \textit{target} word (i.e., one that should \textit{not} be affected by the protected attribute) conditioned on two words describing the protected attribute (e.g., \textit{man} and \textit{woman} in the case of gender). That is, for every target word we seek to minimize: 
\begin{equation}
loss = \sum_{a,b \in \mathcal{P}} p(target|a) - p(target|b)
\end{equation}
where $\mathcal{P} = \{(he,she),(man,woman),(king,queen), \dots \}$ is a set of word pairs characterizing the protected attribute, akin to that used in previous work \cite{DBLP:journals/corr/BolukbasiCZSK16a}.

At this point, the specific form of the objective will depend on the type of word embeddings used. For our expample of SGNS, recall that this algorithm models the conditional probability of a target word given a context word as a function of the inner product of their representations. Though an exact method for calculating the conditional probability includes summing over conditional probability of all the words in the vocabulary, we can use the estimation of log conditional probability proposed by \citet{DBLP:journals/corr/MikolovSCCD13}, i.e., $
\log p(w_O|w_I) \approx \log \sigma ({v'_{wo}}^T v_{wI}) + \sum_{i=1}^{k} [\log {\sigma ({{-v'_{wi}}^T v_{wI}})}]
$.

\paragraph{Nearest Neighbor Bias Mitigation}
Based on observations by \citet{DBLP:journals/corr/abs-1903-03862}, we extend our method to consider the composition of the neighborhood of socially-gendered words of a target word. We note that bias in a word embedding depends not only on the relationship between a target word and explicitly gendered words like \textit{man} and \textit{woman}, but also between a target word and socially-biased male or female words. Bolukbasi et al \cite{DBLP:journals/corr/BolukbasiCZSK16a} proposed a method for eliminating this kind of indirect bias through geometric bias mitigation, but it is shown to be ineffective by the neighborhood metric \cite{DBLP:journals/corr/abs-1903-03862}.


Instead, we extend our method of bias mitigation to account for this neighborhood effect. Specifically, we examine the conditional probabilities of a target word given the $k/2$ nearest neighbors from the male socially-biased words as well as given the $k/2$ female socially-biased words (in sorted order, from smallest to largest). The groups of socially-biased words are constructed as described in the neighborhood metric. If the word is unbiased according to the neighborhood metric, these probabilities should be comparable. We then use the following as our loss function:
\vspace{-0.1cm}
\begin{equation}
    loss = \sum_{i = 0}^{k/2} p(target|m_i) - p(target|f_i),
\end{equation}
where $m$ and $f$ represent the male and female neighbors sorted by distance to the target word $t$ (we use $L1$ distance).

\section{Experiments}


\begin{figure}[htp]
\centering
\includegraphics[width=.33\textwidth]{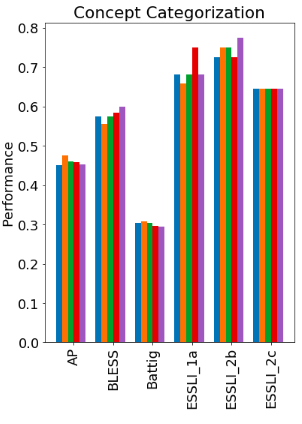}\hfill
\includegraphics[width=.5\textwidth]{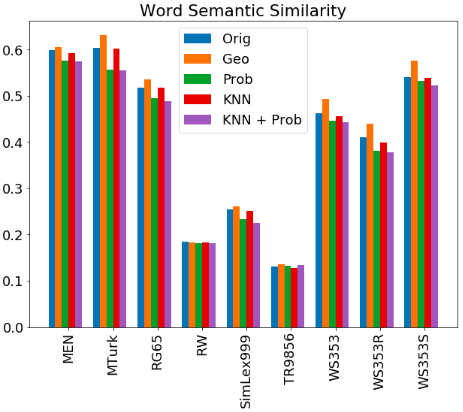}\hfill
\includegraphics[width=.15\textwidth]{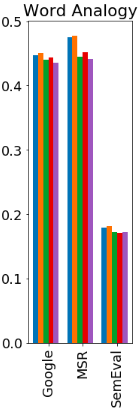}
\caption{Word embedding semantic quality benchmarks for each bias mitigation method (higher is better). See \citet{benchmarks} for details of each metric.}
\label{fig:benchmarks}
\end{figure}

\begin{table}
	\begin{minipage}[b]{0.55\linewidth}
		\centering
  \begin{tabular}{rcc}
  \toprule
                         & RIPA   & Neighborhood         \\
                         \cmidrule(r){2-3} 
                         & $|mean|$ & $|.5 - mean|$ \\     \midrule
    Original             & 2.895  & 0.323       \\
    Geometric            & 0.096  & 0.328       \\
    Simple Probabilistic & 0.320   & 0.250       \\
    Nearest Neighbor     & 1.705  & 0.083       \\
    Composite NN + Prob  & 0.372  & 0.034   \\
    \bottomrule
    \end{tabular}
    \vspace{0.1cm}
		\caption{Remaining Bias (as measured by RIPA and Neighborhood metrics) in fastText embeddings for baseline (top two rows) and our (bottom three) methods.}\label{table:ripa-neighborhood}%
		\label{table:student}
	\end{minipage}\hfill
	\begin{minipage}[b]{0.45\linewidth}
		\centering
		\includegraphics[width=\linewidth]{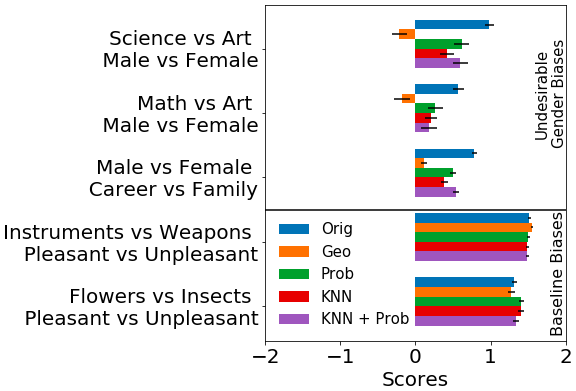}
		\captionof{figure}{Remaining Bias (WEAT score)}\label{fig:remaining-bias}
	\end{minipage}
\end{table}

We evaluate our framework on fastText embeddings trained on Wikipedia (2017), UMBC webbase corpus and statmt.org news dataset (16B tokens) \cite{DBLP:journals/corr/abs-1712-09405}. For simplicity, only the first 22000 words are used in all embeddings, though preliminary results indicate the findings extend to the full corpus. For our novel methods of mitigating bias, a shallow neural network is used to adjust the embedding. The single layer of the model is an embedding layer with weights initialized to those of the original embedding. For the composite method, these weights are initialized to those of the embedding after probabilistic bias mitigation. A batch of word indices is fed into the model, which are then embedded and for which a loss value is calculated, allowing back-propagation to adjust the embeddings. For each of the models, a fixed number of iterations is used to prevent overfitting, which can eventually hurt performance on the embedding benchmarks (See Figure \ref{fig:benchmarks}). We evaluated the embedding after 1000 iterations, and stopped training if performance on a benchmark decreased significantly.

We construct a list of candidate words to debias, taken from the words used in the WEAT gender bias statistics. Words in this list should be gender neutral, and are related to the topics of career, arts, science, math, family and professions (see appendix). We note that this list can easily be expanded to include a greater proportion of words in the corpus. For example, \citet{Ethayarajh2019UnderstandingUW} suggested a method for identifying inappropriately gendered words using unsupervised learning.

We compare this method of bias mitigation with the no bias mitigation ("Orig"), geometric bias mitigation ("Geo"), the two pieces of our method alone ("Prob" and "KNN") and the composite method ("KNN+Prob"). We note that the composite method performs reasonably well according the the RIPA metric, and much better than traditional geometric bias mitigation according to the neighborhood metric, without significant performance loss according to the accepted benchmarks. To our knowledge this is the first bias mitigation method to perform reasonably both on both metrics.

\section{Discussion}
We proposed a simple method of bias mitigation based on this probabilistic notions of fairness, and showed that it leads to promising results in various benchmark bias mitigation tasks. Future work should include considering a more rigorous definition and non-binary of bias and experimenting with various embedding algorithms and network architectures.

\clearpage
\pagebreak

\subsubsection*{Acknowledgements}
The authors would like to thank Tommi Jaakkola for stimulating discussions during the initial stages of this work.

\bibliography{sample-base}

\begin{thebibliography}{12}
\providecommand{\natexlab}[1]{#1}
\providecommand{\url}[1]{\texttt{#1}}
\expandafter\ifx\csname urlstyle\endcsname\relax
  \providecommand{\doi}[1]{doi: #1}\else
  \providecommand{\doi}{doi: \begingroup \urlstyle{rm}\Url}\fi

\bibitem[Bolukbasi et~al.(2016)Bolukbasi, Chang, Zou, Saligrama, and
  Kalai]{DBLP:journals/corr/BolukbasiCZSK16a}
Tolga Bolukbasi, Kai{-}Wei Chang, James~Y. Zou, Venkatesh Saligrama, and Adam
  Kalai.
\newblock Man is to computer programmer as woman is to homemaker? debiasing
  word embeddings.
\newblock \emph{CoRR}, abs/1607.06520, 2016.
\newblock URL \url{http://arxiv.org/abs/1607.06520}.

\bibitem[Islam et~al.(2016)Islam, Bryson, and
  Narayanan]{DBLP:journals/corr/IslamBN16}
Aylin~Caliskan Islam, Joanna~J. Bryson, and Arvind Narayanan.
\newblock Semantics derived automatically from language corpora necessarily
  contain human biases.
\newblock \emph{CoRR}, abs/1608.07187, 2016.
\newblock URL \url{http://arxiv.org/abs/1608.07187}.

\bibitem[Lapowsky(2018)]{googleautocomplete}
Issie Lapowsky.
\newblock Google autocomplete still makes vile suggestions, 2018.
\newblock URL
  \url{"https://www.wired.com/story/google-autocomplete-vile-suggestions/"}.

\bibitem[Anja~Lambrecht(2016)]{lambrecht}
Catherine E.~Tucker Anja~Lambrecht.
\newblock Algorithmic bias? an empirical study into apparent gender-based
  discrimination in the display of stem career ads.
\newblock \emph{SSRN}, 2016.
\newblock \doi{https://papers.ssrn.com/sol3/papers.cfm?abstract_id=2852260}.

\bibitem[Ethayarajh et~al.(2019)Ethayarajh, Duvenaud, and
  Hirst]{Ethayarajh2019UnderstandingUW}
Kawin Ethayarajh, David~Kristjanson Duvenaud, and Graeme Hirst.
\newblock Understanding undesirable word embedding associations.
\newblock In \emph{ACL}, 2019.

\bibitem[Gonen and Goldberg(2019)]{DBLP:journals/corr/abs-1903-03862}
Hila Gonen and Yoav Goldberg.
\newblock Lipstick on a pig: Debiasing methods cover up systematic gender
  biases in word embeddings but do not remove them.
\newblock \emph{CoRR}, abs/1903.03862, 2019.
\newblock URL \url{http://arxiv.org/abs/1903.03862}.

\bibitem[Levy and Goldberg(2014)]{levy2014neural}
Omer Levy and Yoav Goldberg.
\newblock Neural word embedding as implicit matrix factorization.
\newblock In \emph{Advances in neural information processing systems}, pages
  2177--2185, 2014.

\bibitem[Hardt et~al.(2016)Hardt, Price, and
  Srebro]{DBLP:journals/corr/HardtPS16}
Moritz Hardt, Eric Price, and Nathan Srebro.
\newblock Equality of opportunity in supervised learning.
\newblock \emph{CoRR}, abs/1610.02413, 2016.
\newblock URL \url{http://arxiv.org/abs/1610.02413}.

\bibitem[Mikolov et~al.(2013)Mikolov, Sutskever, Chen, Corrado, and
  Dean]{DBLP:journals/corr/MikolovSCCD13}
Tomas Mikolov, Ilya Sutskever, Kai Chen, Greg Corrado, and Jeffrey Dean.
\newblock Distributed representations of words and phrases and their
  compositionality.
\newblock \emph{CoRR}, abs/1310.4546, 2013.
\newblock URL \url{http://arxiv.org/abs/1310.4546}.

\bibitem[Hashimoto et~al.(2016)Hashimoto, Alvarez-Melis, and
  Jaakkola]{hashimoto2016word}
Tatsunori~B Hashimoto, David Alvarez-Melis, and Tommi~S Jaakkola.
\newblock Word embeddings as metric recovery in semantic spaces.
\newblock \emph{Transactions of the Association for Computational Linguistics},
  4:\penalty0 273--286, 2016.

\bibitem[Jastrz{k{e}}bski et~al.(2017)Jastrz{k{e}}bski, Lesniak, and
  Czarnecki]{benchmarks}
Stanis{l}aw Jastrz{k{e}}bski, Damian Lesniak, and Wojciech~Marian Czarnecki.
\newblock How to evaluate word embeddings? on importance of data efficiency and
  simple supervised tasks.
\newblock \emph{CoRR}, abs/1702.02170, 2017.
\newblock URL \url{http://arxiv.org/abs/1702.02170}.

\bibitem[Mikolov et~al.(2017)Mikolov, Grave, Bojanowski, Puhrsch, and
  Joulin]{DBLP:journals/corr/abs-1712-09405}
Tomas Mikolov, Edouard Grave, Piotr Bojanowski, Christian Puhrsch, and Armand
  Joulin.
\newblock Advances in pre-training distributed word representations.
\newblock \emph{CoRR}, abs/1712.09405, 2017.
\newblock URL \url{http://arxiv.org/abs/1712.09405}.

\end{thebibliography}

\appendix

\section{Experiment Notes}

For Equation 4, as described in the original work, in regards to the k sample words $w_i$ is drawn from the corpus using the Unigram distribution raised to the 3/4 power.

For reference, the most male socially-biased words include words such as:’john’, ’jr’, ’mlb’, ’dick’, ’nfl’, ’cfl’, ’sgt’, ’abbot’, ’halfback’, ’jock’, ’mike’, ’joseph’,while the most female socially-biased words include words such as:’feminine’, ’marital’, ’tatiana’, ’pregnancy’, ’eva’, ’pageant’, ’distress’, ’cristina’, ’ida’, ’beauty’, ’sexuality’,’fertility’

\section{Professions}

'accountant',
 'acquaintance',
 'actor',
 'actress',
 'administrator',
 'adventurer',
 'advocate',
 'aide',
 'alderman',
 'ambassador',
 'analyst',
 'anthropologist',
 'archaeologist',
 'archbishop',
 'architect',
 'artist',
 'assassin',
 'astronaut',
 'astronomer',
 'athlete',
 'attorney',
 'author',
 'baker',
 'banker',
 'barber',
 'baron',
 'barrister',
 'bartender',
 'biologist',
 'bishop',
 'bodyguard',
 'boss',
 'boxer',
 'broadcaster',
 'broker',
 'businessman',
 'butcher',
 'butler',
 'captain',
 'caretaker',
 'carpenter',
 'cartoonist',
 'cellist',
 'chancellor',
 'chaplain',
 'character',
 'chef',
 'chemist',
 'choreographer',
 'cinematographer',
 'citizen',
 'cleric',
 'clerk',
 'coach',
 'collector',
 'colonel',
 'columnist',
 'comedian',
 'comic',
 'commander',
 'commentator',
 'commissioner',
 'composer',
 'conductor',
 'confesses',
 'congressman',
 'constable',
 'consultant',
 'cop',
 'correspondent',
 'counselor',
 'critic',
 'crusader',
 'curator',
 'dad',
 'dancer',
 'dean',
 'dentist',
 'deputy',
 'detective',
 'diplomat',
 'director',
 'doctor',
 'drummer',
 'economist',
 'editor',
 'educator',
 'employee',
 'entertainer',
 'entrepreneur',
 'envoy',
 'evangelist',
 'farmer',
 'filmmaker',
 'financier',
 'fisherman',
 'footballer',
 'foreman',
 'gangster',
 'gardener',
 'geologist',
 'goalkeeper',
 'guitarist',
 'headmaster',
 'historian',
 'hooker',
 'illustrator',
 'industrialist',
 'inspector',
 'instructor',
 'inventor',
 'investigator',
 'journalist',
 'judge',
 'jurist',
 'landlord',
 'lawyer',
 'lecturer',
 'legislator',
 'librarian',
 'lieutenant',
 'lyricist',
 'maestro',
 'magician',
 'magistrate',
 'maid',
 'manager',
 'marshal',
 'mathematician',
 'mechanic',
 'midfielder',
 'minister',
 'missionary',
 'monk',
 'musician',
 'nanny',
 'narrator',
 'naturalist',
 'novelist',
 'nun',
 'nurse',
 'observer',
 'officer',
 'organist',
 'painter',
 'pastor',
 'performer',
 'philanthropist',
 'philosopher',
 'photographer',
 'physician',
 'physicist',
 'pianist',
 'planner',
 'playwright',
 'poet',
 'policeman',
 'politician',
 'preacher',
 'president',
 'priest',
 'principal',
 'prisoner',
 'professor',
 'programmer',
 'promoter',
 'proprietor',
 'prosecutor',
 'protagonist',
 'provost',
 'psychiatrist',
 'psychologist',
 'rabbi',
 'ranger',
 'researcher',
 'sailor',
 'saint',
 'salesman',
 'saxophonist',
 'scholar',
 'scientist',
 'screenwriter',
 'sculptor',
 'secretary',
 'senator',
 'sergeant',
 'servant',
 'singer',
 'skipper',
 'sociologist',
 'soldier',
 'solicitor',
 'soloist',
 'sportsman',
 'statesman',
 'steward',
 'student',
 'substitute',
 'superintendent',
 'surgeon',
 'surveyor',
 'swimmer',
 'teacher',
 'technician',
 'teenager',
 'therapist',
 'trader',
 'treasurer',
 'trooper',
 'trumpeter',
 'tutor',
 'tycoon',
 'violinist',
 'vocalist',
 'waiter',
 'waitress',
 'warden',
 'warrior',
 'worker',
 'wrestler',
 'writer'
 
 \section{WEAT Word Sets}
 
 Words used for WEAT statistic, consisting of baseline bias tests and gender bias tests in the format \textbf{X vs Y / A vs B}
 
 \textbf{Flowers vs Insects / Pleasant vs Unpleasant}

\textbf{X:}
"aster",
"clover",
"hyacinth",
"marigold",
"poppy",
"azalea",
"crocus",
"iris",
"orchid",
"rose",
"bluebell",
"daffodil",
"lilac",
"pansy",
"tulip",
"buttercup",
"daisy",
"lily",
"peony",
"violet",
"carnation",
"gladiola",
"magnolia",
"petunia",
"zinnia"
				
\textbf{Y:}
"ant",
"caterpillar",
"flea",
"locust",
"spider",
"bedbug",
"centipede",
"fly",
"maggot",
"tarantula",
"bee",
"cockroach",
"gnat",
"mosquito",
"termite",
"beetle",
"cricket",
"hornet",
"moth",
"wasp",
"blackfly",
"dragonfly",
"horsefly",
"roach",
"weevil"
				
\textbf{A:}
"caress",
"freedom",
"health",
"love",
"peace",
"cheer",
"friend",
"heaven",
"loyal",
"pleasure",
"diamond",
"gentle",
"honest",
"lucky",
"rainbow",
"diploma",
"gift",
"honor",
"miracle",
"sunrise",
"family",
"happy",
"laughter",
"paradise",
"vacation"
				
\textbf{B:}
"abuse",
"crash",
"filth",
"murder",
"sickness",
"accident",
"death",
"grief",
"poison",
"stink",
"assault",
"disaster",
"hatred",
"pollute",
"tragedy",
"divorce",
"jail",
"poverty",
"ugly",
"cancer",
"kill",
"rotten",
"vomit",
"agony",
"prison" 

\textbf{Instruments vs Weapons /  Pleasant vs Unpleasant:} 

\textbf{X:}
"bagpipe",
"cello",
"guitar",
"lute",
"trombone",
"banjo",
"clarinet",
"harmonica",
"mandolin",
"trumpet",
"bassoon",
"drum",
"harp",
"oboe",
"tuba",
"bell",
"fiddle",
"harpsichord",
"piano",
"viola",
"bongo",
"flute",
"horn",
"saxophone",
"violin"

\textbf{Y:}
"arrow",
"club",
"gun",
"missile",
"spear",
"ax",
"dagger",
"harpoon",
"pistol",
"sword",
"blade",
"dynamite",
"hatchet",
"rifle",
"tank",
"bomb",
"firearm",
"knife",
"shotgun",
"teargas",
"cannon",
"grenade",
"mace",
"slingshot",
"whip"

\textbf{A:
}"caress",
"freedom",
"health",
"love",
"peace",
"cheer",
"friend",
"heaven",
"loyal",
"pleasure",
"diamond",
"gentle",
"honest",
"lucky",
"rainbow",
"diploma",
"gift",
"honor",
"miracle",
"sunrise",
"family",
"happy",
"laughter",
"paradise",
"vacation"

\textbf{B:}
"abuse",
"crash",
"filth",
"murder",
"sickness",
"accident",
"death",
"grief",
"poison",
"stink",
"assault",
"disaster",
"hatred",
"pollute",
"tragedy",
"divorce",
"jail",
"poverty",
"ugly",
"cancer",
"kill",
"rotten",
"vomit",
"agony",
"prison" 

\textbf{Male vs Female /  Career vs Family:} 

\textbf{X:
}"brother",
"father",
"uncle",
"grandfather",
"son",
"he",
"his",
"him",
"man",
"himself",
"men",
"husband",
"boy",
"uncle",
"nephew",
"boyfriend",
"king",
"actor"

\textbf{Y:}
"sister",
"mother",
"aunt",
"grandmother",
"daughter",
"she",
"hers",
"her",
"woman",
"herself",
"women",
"wife",
"aunt",
"niece",
"girlfriend",
"queen",
"actress"

\textbf{A:
}"executive",
"management",
"professional",
"corporation",
"salary",
"office",
"business",
"career",
"industry",
"company",
"promotion",
"profession",
"CEO",
"manager",
"coworker",
"entrepreneur"

\textbf{B:}
"home",
"parents",
"children",
"family",
"cousins",
"marriage",
"wedding",
"relatives",
"grandparents",
"grandchildren",
"nurture",
"child",
"toddler",
"infant",
"teenager"

\textbf{Math vs Art / Male vs Female:}

\textbf{X:}
"math",
"algebra",
"geometry",
"calculus",
"equations",
"computation",
"numbers",
"addition",
"trigonometry",
"arithmetic",
"logic",
"proofs",
"multiplication",
"mathematics"

\textbf{Y:}
"poetry",
"art",
"Shakespeare",
"dance",
"literature",
"novel",
"symphony",
"drama",
"orchestra",
"music",
"ballet",
"arts",
"creative",
"sculpture"

\textbf{A:}
"brother",
"father",
"uncle",
"grandfather",
"son",
"he",
"his",
"him",
"man",
"himself",
"men",
"husband",
"boy",
"uncle",
"nephew",
"boyfriend",
"king",
"actor"

\textbf{B:
}"sister",
"mother",
"aunt",
"grandmother",
"daughter",
"she",
"hers",
"her",
"woman",
"herself",
"women",
"wife",
"aunt",
"niece",
"girlfriend",
"queen",
"actress"

\textbf{Science vs Art / Male8 vs Female8:}

\textbf{X:}"science",
"technology",
"physics",
"chemistry",
"Einstein",
"NASA",
"experiment",
"astronomy",
"biology",
"aeronautics",
"mechanics",
"thermodynamics"

\textbf{Y:}
"poetry",
"art",
"Shakespeare",
"dance",
"literature",
"novel",
"symphony",
"drama",
"orchestra",
"music",
"ballet",
"arts",
"creative",
"sculpture"

\textbf{A:}
"brother",
"father",
"uncle",
"grandfather",
"son",
"he",
"his",
"him",
"man",
"himself",
"men",
"husband",
"boy",
"uncle",
"nephew",
"boyfriend"

\textbf{B:}
"sister",
"mother",
"aunt",
"grandmother",
"daughter",
"she",
"hers",
"her",
"woman",
"herself",
"women",
"wife",
"aunt",
"niece",
"girlfriend"



\end{document}